\documentclass{article}

\usepackage{PRIMEarxiv}

\usepackage[utf8]{inputenc} 
\usepackage[T1]{fontenc}    
\usepackage{hyperref}       
\usepackage{url}            
\usepackage{booktabs}       
\usepackage{amsfonts}       
\usepackage{nicefrac}       
\usepackage{microtype}      
\usepackage{fancyhdr}       
\usepackage{graphicx}       
\graphicspath{{media/}}     

\usepackage{amsmath,amssymb,amsfonts}
\usepackage{algorithmic}
\usepackage{textcomp}
\usepackage{xcolor}

\title{TurkEmbed: Turkish Embedding Model on Natural Language Inference \& Sentence Text Similarity Tasks
\thanks{\textit{\underline{Citation}}: 
\textbf{Özay Ezerceli, Gizem Gümüşçekiçci, Tuğba Erkoç, Berke Özenç. "TurkEmbed: Turkish Embedding Model on Natural Language Inference \& Sentence Text Similarity Tasks." 2025 IEEE 11th International Conference on Advances in Software, hardware and Systems Engineering (ASYU), 2025. DOI: 10.1109/ASYU67174.2025.11208511}} 
}

\author{
  Özay Ezerceli \\
  New Mind AI \\
  Istanbul, Türkiye\\
  \texttt{oezerceli@newmind.ai} \\
   \And
  Gizem Gümüşçekiçci, Tuğba Erkoç, Berke Özenç \\
  Faculty of Engineering and Natural Sciences \\
  Isik University \\
  Istanbul, Türkiye\\
  \texttt{\{gizem.gumuscekicci,tugba.erkoc,berke.ozenc\}@isikun.edu.tr} \\
}

\begin{document}
\maketitle

\begin{abstract}
This paper introduces TurkEmbed, a novel Turkish language embedding model designed to outperform existing models, particularly in Natural Language Inference (NLI) and Semantic Textual Similarity (STS) tasks. Current Turkish embedding models often rely on machine-translated datasets, potentially limiting their accuracy and semantic understanding.  TurkEmbed utilizes a combination of diverse datasets and advanced training techniques, including matryoshka representation learning, to achieve more robust and accurate embeddings. This approach enables the model to adapt to various resource-constrained environments, offering faster encoding capabilities. Our evaluation on the Turkish STS-b-TR dataset, using Pearson and Spearman correlation metrics, demonstrates significant improvements in semantic similarity tasks. Furthermore, TurkEmbed surpasses the current state-of-the-art model, Emrecan, on All-NLI-TR and STS-b-TR benchmarks, achieving a 1-4\% improvement. TurkEmbed promises to enhance the Turkish NLP ecosystem by providing a more nuanced understanding of language and facilitating advancements in downstream applications.
\end{abstract}


\section{Introduction}
\label{sec:intro}

Natural Language Processing (NLP) is considered a branch of computational linguistics that focuses on enabling machines to understand, interpret, and generate human language \cite{girgin2022spam}. It has a wide research area and many diverse applications, the most popular ones are sentiment analysis \cite{ezerceli2024mental}, machine translation, and sarcasm detection \cite{girgin2024sentiment}. Many of these applications are popular topics that are currently being investigated to find better approaches to ongoing challenges. With recent advances in technology, the NLP field has seen remarkable advancements and there is a continuous need for improvement.

NLP  applications primarily rely on embeddings to represent words or sentences. An embedding is a numerical representation of words, phrases, or sentences. It is used to convert textual data into numerical vector representations that preserve semantic and syntactic properties. Thus, the performance of embedding systems plays a crucial role in determining the success of NLP systems.

Word embeddings are categorized as non-contextual and contextual, each with distinct impacts on NLP models like TurkEmbed. Non-contextual embeddings, such as Word2Vec \cite{mikolov2013efficientestimationwordrepresentations}, GloVe \cite{pennington-etal-2014-glove}, and FastText \cite{bojanowski2017enrichingwordvectorssubword}, assign fixed representations to words, ignoring variations in meaning based on context \cite{almeida2023wordembeddingssurvey}. For instance, the word "bank" is represented identically whether it refers to a "riverbank" or a "financial institution." This limitation makes non-contextual embeddings inadequate for advanced tasks like Natural Language Inference (NLI) and Semantic Textual Similarity (STS), which require contextual understanding \cite{miaschi-dellorletta-2020-contextual}.

Contextual embeddings, such as BERT \cite{devlin2019bertpretrainingdeepbidirectional}, ELMo \cite{peters2018deepcontextualizedwordrepresentations}, and T5 \cite{raffel2023exploringlimitstransferlearning}, address this by generating dynamic representations that adapt to a word’s usage within a sentence, enabling nuanced interpretation. This is crucial for Turkish, a morphologically rich language. TurkEmbed leverages techniques like matryoshka representation learning \cite{kusupati2022matryoshka} to optimize contextual embeddings, overcoming challenges in Turkish morphology and syntax. These advancements position TurkEmbed as a state-of-the-art model for Turkish NLP, highlighting the importance of contextual embeddings in capturing semantic nuances.

Embedding models are often language-dependent, with multilingual versions available but potentially less effective for low-resource languages like Turkish due to the need for generalization across multiple languages. While most embedding models are built for English due to the abundance of resources, Turkish NLP research relies on multilingual models or a limited number of Turkish-specific models, which can hinder performance in tasks requiring a deep understanding of semantic relationships; even models like bert-base-turkish-cased-mean-nli-stsb-tr \cite{emrecan2024bert} have room for improvement in semantic similarity tasks.

We introduce TurkEmbed, a novel and enhanced Turkish embedding model designed to overcome existing limitations in the Turkish language. The methodology involves combining diverse datasets with advanced training techniques, notably Matryoshka representation learning, and selecting base models from the MTEB leaderboard \cite{muennighoff2023mtebmassivetextembedding}. TurkEmbed's performance was evaluated on semantic similarity tasks using the Turkish STSb \cite{beken-fikri-etal-2021-semantic} and STS22 \cite{sts22dataset} datasets, showing superior results compared to current state-of-the-art models. The main contributions include the TurkEmbed model itself, which excels on NLI and STS tasks, the demonstration of Matryoshka learning's efficacy for Turkish, and a thorough evaluation on benchmarks like All-NLI-TR, STSb-TR, and STS22-Crosslingual-STS, ultimately aiming to advance Turkish NLP capabilities.


The remainder of this paper is organized as follows: Section 2 reviews related works on Turkish embedding models. Section 3 details the methodology employed in developing TurkEmbed. Section 4 describes the experimental setup, and Section 5 presents the results and discussion. Finally, Section 6 concludes the paper and outlines future research directions.

\section{Related Work}

The Turkish language, characterized by its agglutinative morphology and rich vocabulary, presents a unique and compelling challenge for natural language processing (NLP).  The development of effective word embedding models is paramount to overcoming these linguistic complexities and enabling robust NLP tools for Turkish.

Contextual embedding models like BERT \cite{vaswani2023attentionneed} and ELMo \cite{peters2018deepcontextualizedwordrepresentations} significantly changed NLP by overcoming the limitations of non-contextual approaches. ELMo employed deep bidirectional language models for context-aware representations, whereas BERT used the Transformer architecture with bidirectional encoders to achieve state-of-the-art results across many NLP tasks. BERT's capacity to consider the full context of a word was particularly advantageous for Turkish, facilitating a more nuanced understanding of meaning in complex sentences. Research confirmed that BERT-based models outperformed earlier non-contextual methods in various Turkish NLP tasks, highlighting the crucial role of contextual awareness for the language.

Building upon the success of BERT, Turkish-specific BERT models are developed and trained on large-scale Turkish corpora like the Boun Web Corpus \cite{sak:08} and the Huawei Corpus \cite{huaweicorpus}. These models, pre-trained on extensive Turkish text, offered enhanced performance on downstream tasks due to a better understanding of language-specific nuances.  Furthermore, research has explored adapting and fine-tuning these models for specific Turkish NLP tasks. \cite{emrecan2024bert}'s work, for example, utilized machine-translated datasets to fine-tune BERT-based models for Turkish NLI and STS, establishing initial benchmarks for these tasks and highlighting the feasibility of cross-lingual transfer learning for Turkish NLP. The YTU Cosmos Lab \cite{kesgin2023developing} further introduced Turkish BERT models trained on a sizable corpus of 75GB, compiled from various sources to enhance the diversity and representativeness of the data. These models aimed to improve performance on downstream tasks such as text classification and named entity recognition. Although they provided a solid foundation, they required significant computational resources for training and did not specifically address tasks like NLI and STS.

Beyond core NLP tasks, adaptations of existing architectures have also emerged.  Turkish-ColBERT \cite{santhanam2021colbertv2} adapted the ColBERT architecture, initially designed for English information retrieval, to the Turkish language, showcasing the adaptability of advanced retrieval models to morphologically complex languages. It was fine-tuned on the machine-translated MS MARCO dataset \cite{nguyen2016msmarco}, utilizing over 500,000 translated queries and passages. While it demonstrated improved retrieval performance, the reliance on machine-translated data posed challenges in capturing idiomatic expressions and linguistic nuances unique to Turkish.  

In addition, the exploration of multilingual models such as XLM-ROBERTa \cite{conneau2019unsupervised} and multilingual E5 \cite{wang2024multilingual} has opened the doors for the learning of cross-lingual transfer in Turkish NLP. These models, trained on data from more than 100 languages, leverage shared representations to improve performance in low-resource languages such as Turkish, offering a cost-effective approach to building robust Turkish NLP systems.  Similarly, the GTE-multilingual-base model \cite{zhang2024mgte}, providing generalized embeddings suitable for various tasks across multiple languages, offers a versatile solution for Turkish NLP, particularly in cross-lingual applications.

In addition, general embedding models show promise for Turkish NLP. One such model is nomic-ai/nomic-embed-text-v2-moe \cite{nussbaum2025trainingsparsemixtureexperts}, a state-of-the-art multilingual text embedding model using a Mixture of Experts (MoE) architecture. Trained on over 1.6 billion data pairs across approximately 100 languages, including Turkish, it offers competitive performance for multilingual retrieval tasks, making it efficient for Turkish applications requiring cross-language capabilities.

Another significant model is Alibaba-NLP/gte Modernbert-base \cite{zhang2024mgte}, part of their GTE series. It is a ModernBERT base model language support is English. The GTE series includes models supporting a wide range, potentially including Turkish, with multilingual variants designed for long context lengths and trained on diverse datasets. These models suggest that, with fine-tuning or adaptation, they could further enhance Turkish NLP tasks, especially those involving longer texts or cross-language comparisons, opening exciting avenues for future research.

\section{Methodology}
\subsection{Model Selection}

A comprehensive selection process was undertaken to identify appropriate base models for subsequent fine-tuning on Turkish language tasks, considering leading native English, Turkish, and multilingual candidates known for capturing language-specific nuances. The selection criteria prioritized models with demonstrated effectiveness in cross-lingual transfer learning and semantic understanding capabilities, particularly for morphologically rich languages like Turkish.

Among the evaluated models were ModernBERT-base (150M parameters) and its larger variant ModernBERT-large (396M) \cite{warner2024smarter}, representing state-of-the-art encoder architectures with improved efficiency; the instruction-tuned KaLM-embedding-multilingual-mini-instruct-v1.5 (494M) \cite{hu2025kalm}, specifically designed for multilingual embedding tasks; the compact paraphrase-multilingual-MiniLM-L12-v2 (118M) suitable for resource-constrained scenarios \cite{reimers2019sentence}; the generalized multilingual text embedding models GTE-multilingual-base (305M) and gte-modernbert-base (149M) from Alibaba's Tongyi Lab \cite{zhang2024mgte}, known for their strong multilingual capabilities; the XLM-RoBERTa-based multilingual-E5-large-instruct (560M) \cite{wang2024multilingual}, which has demonstrated superior performance on various multilingual benchmarks; and the multilingual Mixture-of-Experts model, nomic-embed-text-v2-moe \cite{nussbaum2025trainingsparsemixtureexperts}, offering scalable parameter efficiency.

The model selection process considered computational efficiency, multilingual capabilities, and proven performance on semantic similarity tasks. Models with strong cross-lingual transfer capabilities were prioritized, given the limited availability of high-quality Turkish training data compared to resource-rich languages like English.

\subsection{Loss Functions and Their Theoretical Foundations}

The selection of loss functions for our two-stage training pipeline was guided by both theoretical considerations and empirical evidence from recent advances in sentence embedding research. Multiple Negatives Ranking Loss \cite{gao2021scaling}, employed with the All-NLI-TR dataset, was selected for the initial training stage due to its contrastive learning efficiency that leverages in-batch negatives, treating all other sentences in the batch as negative examples for each anchor-positive pair. This approach enables efficient learning from a large number of negative examples without requiring explicit negative sampling strategies \cite{karpukhin2020dense}, and its contrastive nature aligns naturally with NLI data structure where positive pairs (entailment relationships) must be distinguished from negative pairs \cite{reimers2019sentence}.

CoSENT Loss \cite{su2022cosent}, applied to the STSB-TR dataset in the second training stage, was chosen for its training-inference consistency that directly optimizes cosine similarity between sentence embeddings, creating perfect alignment between the training objective and the inference-time similarity metric. This consistency reduces the gap between training and deployment while providing smoother gradients across the full range of similarity scores, allowing for more stable training and better convergence on continuous similarity prediction tasks \cite{gao2021simcse}. Research has demonstrated that CoSENT Loss produces better-calibrated similarity scores that align more closely with human similarity judgments \cite{su2022cosent}.

Matryoshka Loss \cite{kusupati2022matryoshka} integrates with both primary loss functions, enabling the model to learn embeddings across multiple dimensions concurrently. This integration provides adaptive deployment capabilities allowing a single model to generate useful embeddings at various dimensions (64, 128, 256, 512, 768), dimension efficiency where even truncated embeddings maintain competitive performance, and resource optimization that reduces computational and storage requirements for production environments \cite{li20242d}.


\subsection{Training Procedure and Sequential Learning Rationale}
 
\begin{figure}
    \centering
    \includegraphics[width=0.4\linewidth]{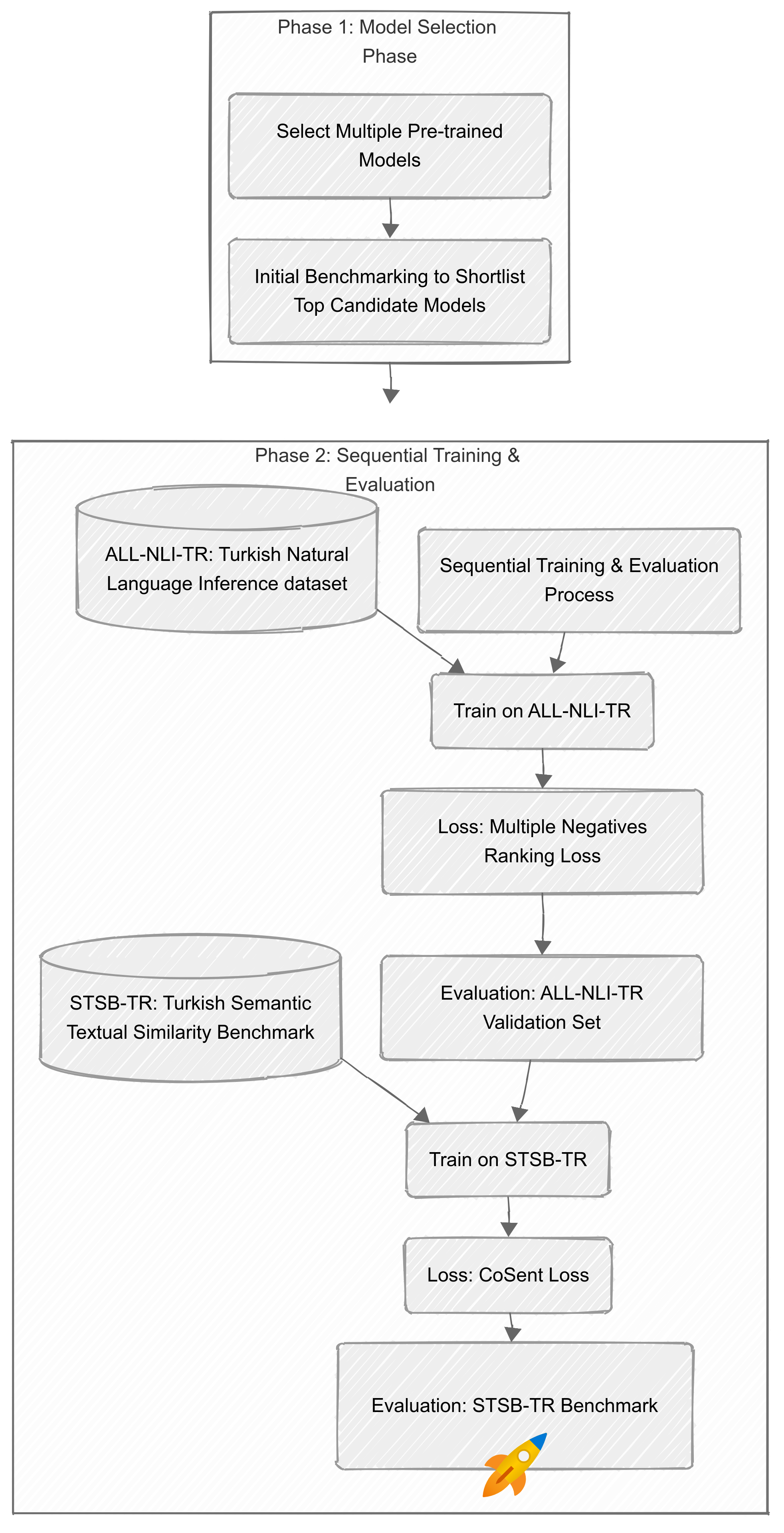}
    \caption{TurkEmbed Sequential Training Pipeline }
    \label{fig:turkembed-flowchart}
\end{figure}

Our training methodology employs a carefully designed two-stage sequential fine-tuning process, illustrated in Figure \ref{fig:turkembed-flowchart}, addressing the specific challenges of developing high-quality Turkish language embeddings. The All-NLI-TR dataset was selected as the initial training corpus because NLI datasets provide explicit semantic relationships (entailment, contradiction, neutral) that force models to learn fine-grained semantic distinctions essential for high-quality embeddings \cite{conneau2017supervised}. Multiple studies have demonstrated that NLI training creates robust sentence representations that transfer effectively to various semantic tasks, including STS, developing structural knowledge about semantic relationships that generalizes well across domains \cite{conneau2017supervised, wang2018glue}. Additionally, NLI training improves cross-lingual transfer capabilities, particularly valuable for Turkish as a morphologically rich language with limited resources, and provides superior zero-shot transfer capabilities compared to models trained directly on specific downstream objectives \cite{artetxe2019cross, yang2018learning}.

The decision to follow NLI training with STSB-TR fine-tuning is supported by substantial empirical evidence showing that while NLI focuses on categorical relationships between sentences, STS provides continuous similarity scores that help models refine their understanding of semantic similarity in a more nuanced way. Sequential fine-tuning on STSB-TR after NLI training helps prevent catastrophic forgetting of semantic distinctions learned in the first stage while enabling task-specific specialization, consistently outperforming simultaneous multi-task training and other training regimens \cite{ruder2017overview, reimers2019sentence}.

The training process begins with model initialization and sequence length adjustment, followed by first-stage fine-tuning on All-NLI-TR using Multiple Negatives Ranking Loss within the Matryoshka loss framework, with validation performed on All-NLI-TR and testing on STSB-TR. The second stage continues with STSB-TR fine-tuning using CoSENT Loss combined with Matryoshka loss to enhance sentence similarity capabilities. Final evaluation employs Triplet and Embedding Similarity Evaluators across various embedding dimensions \cite{reimers-2019-sentence-bert}. Training durations varied according to model size and complexity, with All-NLI-TR training requiring approximately 30 minutes to 1 hour and STSB-TR fine-tuning taking 8 to 30 minutes. Smaller models like TurkEmbed (305M parameters) benefit from shorter training durations compared to larger models like multilingual-E5-large-instruct (560M parameters), demonstrating the efficiency and scalability of our approach.



\section{Experiments}
\subsection{Experimental Setup}

The experiments were conducted on a high-performance computing setup equipped with an NVIDIA A100 40GB GPU. Python 3.11.11, PyTorch 2.5.1+cu121, and Transformers 4.49.0.dev0, complemented by Sentence Transformers 3.3.1 and Datasets 3.2.0. HuggingFace’s Transformers and Datasets libraries were utilized for efficient model handling and dataset loading, ensuring seamless integration and scalability.

The hyperparameters were carefully selected to optimize model performance while preventing overfitting. Batch sizes of 16, 32, 64, and 128 were employed, depending on the model and GPU memory constraints. The learning rate was tuned within the range of $1 \times 10^{-5}$ to $8 \times 10^{-5}$, and the number of training epochs was set between 1 and 10. To enhance training stability, a warmup ratio of 0.1 or specific warmup steps based on dataset size was applied. The maximum sequence length was adjusted to 75, 128, 256, or 512 tokens, depending on the model’s capacity and task requirements, ensuring efficient processing of input data.

Three key metrics were used to assess the performance of the model. The Pearson Correlation Coefficient measured the linear correlation between predicted and actual similarity scores, providing insights into the model's ability to capture semantic relationships. Spearman’s Rank Correlation Coefficient evaluated the monotonic relationship between predicted and actual rankings, ensuring robustness in capturing relative similarities. Additionally, accuracy was used for Natural Language Inference (NLI) tasks to determine the percentage of correct predictions, offering a comprehensive assessment of the model's overall effectiveness. These metrics collectively ensured a rigorous evaluation of TurkEmbed’s performance in various tasks and datasets.

\subsection{Datasets}

\textbf{The All-NLI-TR dataset} is a combination of the SNLI \cite{bowman2015large} and MultiNLI \cite{williams2017broad} datasets translated into Turkish. It contains 482,091 training samples, 6,802 for development, and 6,827 for testing, covering a diverse range of genres and topics. The dataset includes pairs of sentences labeled with entailment, contradiction, or neutral, providing a robust foundation for training models on NLI tasks.

\textbf{The STSB-TR dataset} \cite{hercig2021evaluation} is a Turkish version of the Semantic Textual Similarity Benchmark, containing sentence pairs with similarity scores ranging from 0 to 5. It includes 5,749 training samples, 1,500 validation samples, and 1,379 test samples. This dataset enables models to learn fine-grained semantic relationships between sentences.

\section{Results and Discussion}

\begin{table}[h!]
    \scriptsize
    \centering
    \caption{Performance on All-NLI-TR Test Set}
    \label{table:1}
    \begin{tabular}{|l|l|l|l|}
        \hline
        \textbf{Model} & \textbf{Max Seq} & \textbf{Embedding} & \textbf{Cosine Accuracy} \\
         & \textbf{Length}  & \textbf{Dimension} & \\ \hline
        gte-multilingual-base & 8192 & 768 & 0.896   \\ \hline
        bge-m3 & 8192 & 1024 &  0.914   \\ \hline
        turkish-e5-large & 514 & 1024 &  0.876   \\ \hline
        Qwen3-Embedding-8B & 32000 & 32 to 4096 &  0.876   \\ \hline
        ModernBERT-base & 8192 & 768 & 0.605 \\ \hline
        ModernBERT-large & 8192 & 1024 & 0.601 \\ \hline
        KaLM-embedding-multi & 131072 & 896 & 0.864 \\
        lingual-mini-instruct-v1.5 &     &     &   \\  \hline
        paraphrase-multilingual & 512 & 384 & 0.902   \\
        -MiniLM-L12-v2 &  &   &   \\ \hline
        multilingual-E5-large & 514 & 1024 & 0.881 \\
        -instruct &  &   &   \\ \hline
        nomic-embed- & 2048 & 256 to 768 & 0.821   \\
        text-v2-moe &  & &   \\ \hline
        Emrecan's Model & 512 & 768 & 0.885 \\ \hline
        TurkEmbed-All-NLI-TR  & 8192 & 64 to 768 & \textbf{0.935} \\ \hline
        TurkEmbed4STS  & 8192 & 64 to 768 & \underline{0.924} \\\hline
        modernbert-base-tr-uncased & 8192 & 256 to 768 & \underline{0.924} \\
        -allnli-stsb & & & \\
        
        \hline
    \end{tabular}
\end{table}

\subsection{Performance on All Natural Language Inference} 

To rigorously assess the model's resilience to catastrophic forgetting following sequential training, TurkEmbed's performance was evaluated on the All-NLI-TR test set subsequent to its fine-tuning on the STSb-TR dataset. The results, presented in Table \ref{table:1}, indicate that TurkEmbed achieved superior performance compared to all other evaluated models, obtaining a cosine accuracy of 0.935 before stsb fine-tuning and 0.924 after fine-tuning. Notably, this surpasses the performance of strong contemporary models such as bge-m3 and paraphrase-multilingual-MiniLM-L12-v2 under the same evaluation conditions. This outcome suggests that the proposed training methodology incorporating Matryoshka representation learning effectively mitigates catastrophic forgetting, yielding a robust and versatile embedding model capable of retaining task-specific knowledge across different training phases within the Turkish NLP context.

\subsection{Performance on Semantic Textual Similarity Benchmark}

\begin{table}[h!]
    \scriptsize
    \centering
    \caption{Performance on STSB-TR Test Set}
    \label{tab:stsb-tr-performance} 
    \begin{tabular}{|l|l|l|l|l|}
        \hline
        \textbf{Model} & \textbf{Max Seq} & \textbf{Embedding} & \textbf{Pearson} & \textbf{Spearman} \\
         & \textbf{Length}  & \textbf{Dimension} & \textbf{Cosine} & \textbf{Cosine} \\
        \hline
        gte-multilingual-base & 8192 & 768 & 0.804 & 0.804  \\ \hline
        bge-m3 & 8192 & 1024 & 0.795 & 0.797 \\ \hline
        turkish-e5-large & 514 & 1024 &  0.795 & 0.800   \\ \hline
        Qwen3-Embedding-8B & 32000 & 32 to 4096 &  0.798 & 0.794 \\ \hline
        ModernBERT-base & 8192 & 768 & 0.758 & 0.749   \\ \hline
        ModernBERT-large & 8192 & 1024 & 0.772 & 0.771 \\ \hline
        KaLM-embedding-multi & 131072 & 896 & 0.797 & 0.802 \\
        lingual-mini-instruct-v1.5 &  &   &   &   \\ \hline
        paraphrase-multilingual & 512 & 384 & 0.812 & 0.825 \\
        -MiniLM-L12-v2 &  &   &   &   \\ \hline
        multilingual-E5-large & 514 & 1024 & \textbf{0.846} & \textbf{0.854} \\
        -instruct &  &   &   &   \\ \hline
        nomic-embed-text-v2-moe & 2048 & 768 & 0.828 & \underline{0.834} \\ \hline
        Emrecan's Model & 512 & 768 & \underline{0.834} & 0.830 \\ \hline
        TurkEmbed-All-NLI-TR  & 8192 & 64 to 768 & 0.813 & 0.820 \\ \hline
        TurkEmbed4STS  & 8192 & 64 to 768 & \textbf{0.845} & \textbf{0.853} \\ \hline
        modernbert-base-tr-uncased & 8192 & 256 to 768 & 0.825 & 0.832 \\
        -allnli-stsb & & & & \\
        \hline
    \end{tabular}
\end{table}

In the final fine-tuning on STSB-TR, TurkEmbed achieved state-of-the-art results for Turkish semantic tasks, with Pearson and Spearman correlations of 0.845 and 0.853, given in Table \ref{tab:stsb-tr-performance}. It closely rivals multilingual-E5-large-instruct while using nearly half the parameters (305M vs. 560M), making it more efficient. It also outperforms models like nomic-embed-text-v2-moe and Emrecan’s Model. Models with poor Turkish performance, such as gte-modernBERT-base and IBM Granite, were excluded. TurkEmbed’s strong accuracy and efficiency make it a top choice for Turkish NLP.

\subsection{Evaluation on STS22-cross-lingual Semantic Textual Similarity (TR Subset)}

To evaluate generalization capabilities, TurkEmbed4STS was assessed on the Turkish STS subset derived from the STS22-Crosslingual-STS dataset. The model achieved a Pearson cosine correlation of 0.646 and a Spearman cosine correlation of 0.668, as presented in the Table \ref{table:3}. These results position TurkEmbed4STS competitively, surpassing several models, including Emrecan's Model and demonstrating performance comparable to the top-performing nomic-embed-text-v2-moe. Qwen3-Embedding-8B achieved the highest Pearson and Spearman cosine correlations at 0.701 and 0.721, respectively.

\begin{table}[h!]
    \scriptsize
    \centering
    \caption{Performance on STS22-Crosslingual-STS}
    \label{table:3}
    \begin{tabular}{|l|l|l|}
        \hline
        \textbf{Model} & \textbf{Pearson} & \textbf{Spearman} \\
        \textbf{} & \textbf{Cosine} & \textbf{Cosine} \\
        \hline
        gte-multilingual-base & 0.647 & 0.669   \\ \hline
        bge-m3 & 0.663 & 0.698 \\ \hline
        turkish-e5-large & 0.668 & 0.692 \\ \hline
        multilingual-E5-large & \underline{0.676}  & 0.695   \\
        -instruct &  & \\ \hline
        Qwen3-Embedding-8B & \textbf{0.701} & \textbf{0.721} \\ \hline
        ModernBERT-base & 0.436 & 0.471 \\ \hline
        ModernBERT-large & 0.375 & 0.380 \\ \hline
        KaLM-embedding-multi & 0.342 & 0.365 \\ 
        lingual-mini-instruct-v1.5 &  & \\ \hline
        Emrecan's Model & 0.540 & 0.563 \\ \hline
        NeoBERT & 0.622 & 0.663 \\ \hline
        nomic-embed- & 0.653 & \underline{0.706} \\
        text-v2-moe &  &   \\ \hline
        Emrecan's Model & 0.540 & 0.563 \\ \hline
        TurkEmbed4STS & 0.646 & 0.668 \\\hline
        modernbert-base-tr-uncased-allnli-stsb & 0.520 & 0.559 \\
        \hline
    \end{tabular}
\end{table}

\subsection{Inference Speed Comparison}

Inference speed is a critical factor for real-world applications, especially for tasks requiring real-time processing. TurkEmbed's inference speed was compared with Emrecan's model on the Google Colab T4 GPU using a batch size of 32 and 10,000 samples from the All-NLI-TR dataset. For tensor type FP32, TurkEmbed's encoding speed was approximately 310 sentences per second, which is 2.17 times slower than Emrecan's model. For tensor type FP16, TurkEmbed achieved an encoding speed of 1,561 sentences per second, 1.23 times slower than Emrecan's model, as given in Table \ref{table:4}. This difference in speed is largely attributed to TurkEmbed's larger size, as it has nearly three times the parameters of Emrecan's model. Despite the slower speed, TurkEmbed’s advanced architecture and higher accuracy make it a valuable choice for applications prioritizing performance over speed.

\begin{table}[h!]
    \scriptsize
    \centering
    \caption{Inference Speed Comparison}
    \label{table:4}
    \begin{tabular}{|l|l|l|}
        \hline
        \textbf{Model} & \textbf{Inference Speed} & \textbf{Tensor} \\
        \textbf{} & \textbf{(sentences/sec)} & \textbf{Type} \\
        \hline
        Emrecan's Model & $\sim$675 & FP32   \\ \hline
        Emrecan's Model & $\sim$1933 & FP16 \\ \hline
        TurkEmbed & $\sim$310 & FP32 \\ \hline
        TurkEmbed & $\sim$1561 & FP16 \\
        \hline
    \end{tabular}
\end{table}

\section{Conclusion}

This paper introduced TurkEmbed, a novel Turkish embedding model addressing the limitations of existing approaches, particularly the reliance on machine-translated datasets and difficulties capturing Turkish morphology. By fine-tuning strong multilingual base models (gte-multilingual-base, multilingual-e5 large-instruct) with advanced techniques, including Matryoshka representation learning, TurkEmbed achieves state-of-the-art performance on Turkish NLI (ALL-NLI-TR) and STS (STSB-TR) benchmarks, evaluated using Pearson and Spearman correlations. These results demonstrate the effectiveness of adapting multilingual models to enhance language specificity for resource-limited, morphologically rich languages. Future research will explore larger model architectures, integration of native Turkish datasets, transfer learning opportunities, and the evaluation of TurkEmbed across diverse downstream applications and real-world deployment scenarios.

\bibliographystyle{unsrt}  
\bibliography{references}  


\end{document}